\documentclass[conference]{IEEEtran}

\usepackage{cite}
\usepackage{amsmath,amssymb,amsfonts}
\usepackage{algorithmic}
\usepackage{graphicx}
\usepackage{textcomp}
\usepackage{booktabs}
\usepackage{xcolor}

\def\BibTeX{{\rm B\kern-.05em{\sc i\kern-.025em b}\kern-.08em
    T\kern-.1667em\lower.7ex\hbox{E}\kern-.125emX}}

\begin{document}
%

\bibliographystyle{IEEEtran}


\title{Perceptual Asymmetry Between Hue Categories: Evidence from Human Color Categorization}





\author{
\IEEEauthorblockN{Elnara Kadyrgali, Nuray Toganas, Muragul Muratbekova, Pakizar Shamoi\IEEEauthorrefmark{1}}
\IEEEauthorblockA{School of Information Technology and Engineering\\
Kazakh-British Technical University,
Almaty, Kazakhstan 050000\\ 
Email: p.shamoi@kbtu.kz\IEEEauthorrefmark{1}
}
}



\maketitle

\begin{abstract}

Human color categories are not uniformly distributed in perceptual space, yet most computational color models still assume fixed and evenly structured representations. In this paper, we present a focused analytical extension of the COLIBRI fuzzy color model by investigating perceptual asymmetry between hue categories. Using previously collected large-scale human color categorization data, we introduce quantitative measures of category extent and boundary uncertainty, namely Wideness and Boundary Width, derived from fuzzy membership functions at the $\alpha = 0.5$ level. The analysis reveals a strong imbalance between the two categories: yellow occupies a compact and sharply constrained region of the hue space, whereas green spans a substantially broader interval and exhibits a more extended transition structure. The results show that perceptual color categories are not only fuzzy, but also highly non-uniform in their geometric organization. This asymmetry suggests that some categories behave as narrow, highly specific perceptual labels, while others function as broad, tolerant regions of human color naming. These findings provide a new perspective on linguistic color categorization and extend the interpretability of the COLIBRI framework for perceptually grounded color modeling.


\end{abstract}

\IEEEpeerreviewmaketitle

\section{Introduction}
Colors play a fundamental role in human perception and interaction with the environment \cite{Elliot2014}. Despite significant advances in color representation, existing computational models still struggle to accurately reflect how humans perceive colors. Conventional color spaces, such as RGB, HSV, and CIE Lab, rely on precise numerical definitions and fixed partitions, which do not account for the inherent uncertainty and linguistic nature of human color perception \cite{kim2024fuzzycolormodelclustering, burambekova2024comparative}.

Humans typically describe colors using natural language, where category boundaries are gradual, overlapping, and context-dependent \cite{Lindsey2021, Hansen2017}. This discrepancy between numerical representation and perceptual interpretation limits the effectiveness of color-based systems in applications such as image retrieval, design, and human--computer interaction.

\begin{figure*}
  \centering
  \includegraphics[width=\textwidth]{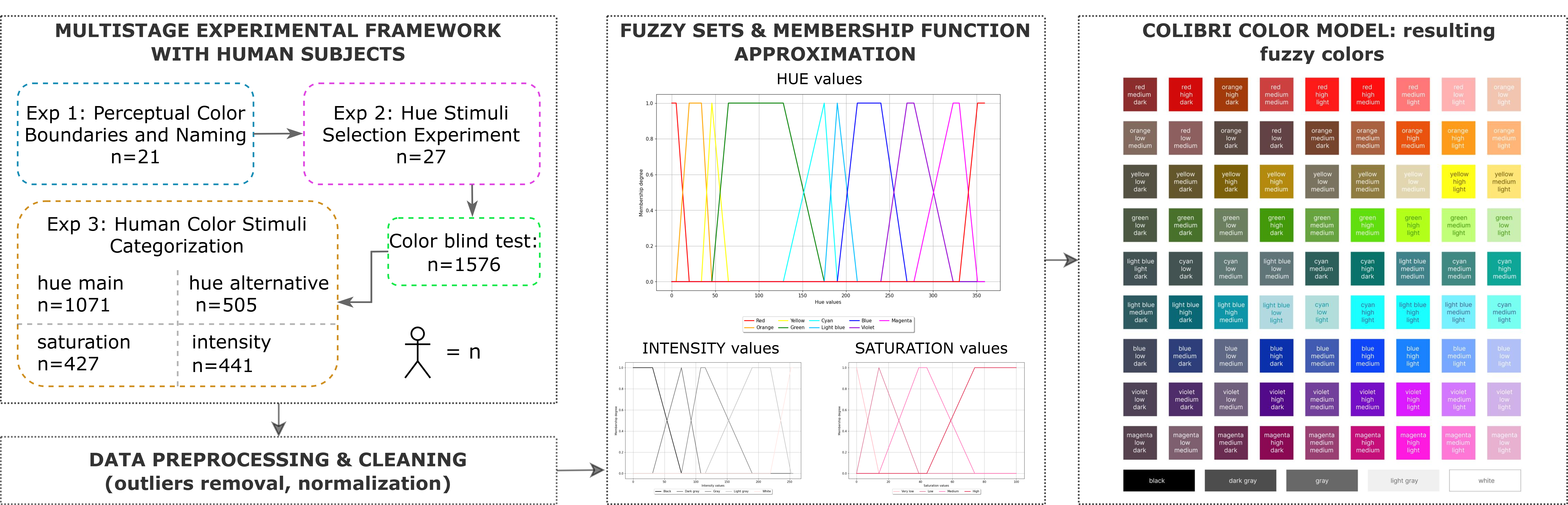}
  \caption{Overview of the COLIBRI Fuzzy Color Modeling Framework} 
  \label{colibri}
\end{figure*}

To address this gap, the Human Perception-Based Fuzzy Color Model, COLIBRI (Color Linguistic-Based Representation and Interpretation), was previously introduced as a framework that integrates fuzzy set theory with linguistically grounded color categories (see Fig. \ref{colibri}) \cite{colibriAccess}. We constructed the model based on a large-scale experimental study with 2,496 participants, which enabled us to derive perceptually meaningful fuzzy partitions of hue, saturation, and intensity. It captures the ambiguity and overlap inherent in human color categorization by allowing colors to belong to multiple categories with varying degrees of membership, and has been validated through statistical analysis and practical applications such as image labeling for context-based image retrieval.

In this paper, we extend the original COLIBRI study by presenting additional analysis and results derived from the same large-scale dataset that were not included in the initial publication. Specifically, we provide a deeper investigation of perceptual patterns, extended statistical evaluations, and further insights into the structure and variability of linguistic color categories. 

In our previous work \cite{colibriAccess}, we demonstrated that hue categories are perceptually non-uniform, with some colors occupying substantially wider fuzzy regions than others. However, this observation was not examined as a focused research problem in its own. 
Therefore, the goal of this paper is to provide a targeted comparative analysis of perceptually asymmetric hue categories.

The paper has the following structure. Section I is this Introduction. Section II reviews related work on human color categorization and perceptual variability. 
Section III introduces the proposed methodology, including the definition of the Wideness and Boundary Width metrics for analyzing fuzzy color categories. 
Section IV presents the experimental results and a comparative analysis of hue categories using the COLIBRI dataset. 
Section V discusses the implications of the observed perceptual asymmetry from perceptual, linguistic, and methodological perspectives. 
Finally, Section VI concludes the paper and suggests directions for future work.

\section{Related Work}

Human color categorization is characterized by significant inter-observer variability and non-uniform category boundaries. Multiple studies report that color naming varies systematically across individuals, and category boundaries behave largely independently rather than forming a single consistent partition of color space \cite{EMERY201766, fairchild2013metameric}.  This variability is most evident in boundary regions, where colors are ambiguously categorized and show low interobserver agreement \cite{Brasil2014ColorNT, parraga2009psychophysical}.

Furthermore, color categorization depends on task conditions and stimulus properties. Studies show that category effects can vary depending on how the task is performed and, in some cases, may appear weak or inconsistent. In addition, certain regions of the color space, especially transitional and low saturation colors, tend to show lower agreement between observers and higher uncertainty in labeling \cite{Witzel2016CategoricalPF, lindsey2006universality}. These results suggest that color categories are not fixed, but depend on context and individual perception.

To better understand how such uncertainty is captured, existing studies measure uncertainty in color categorization using behavioral indicators such as naming variability and confidence \cite{Alvarado2005, EMERY201766, Hsieh2020}, as well as statistical and probabilistic approaches including entropy and probabilistic modeling \cite{Min_Park2026,McMahan2015, Hardman_2016}. While these approaches capture disagreement and variability in human responses, they primarily rely on response-based or probabilistic representations. As a result, they do not provide a direct geometric interpretation of category structure in the color space.

These limitations highlight the need for models that can explicitly represent uncertainty, variability, and overlap in color categorization. In this context, fuzzy color models represent colors through membership functions over linguistic categories, enabling soft assignment across multiple classes. In data-driven approaches, these functions are estimated from human responses rather than predefined analytically.

Within this framework, COLIBRI constructs fuzzy partitions for hue, saturation, and intensity using a three-stage pipeline: (i) elicitation of perceptual category boundaries and linguistic labels, (ii) refinement of distinguishable stimuli within each attribute, and (iii) large-scale categorization used to estimate membership functions from response distributions.

This procedure enables direct modeling of category overlap, non-uniform widths, and boundary uncertainty, while representing each color as a vector of membership degrees derived from empirical data.

\section{Methodology}

Based on the fuzzy set modelling of hue values, we consider a set of nine primary hue categories - \textit{red, orange, yellow, green, cyan, light blue, blue, violet, magenta}:

\begin{equation}
    {X} = \{x_1, \dots, x_9\} = \{\text{red, orange, \dots, magenta}\}
\end{equation}

Each category $x_i$ is represented by a fuzzy membership function:

\begin{equation}
    \mu_i(h): [0, 360^\circ] \rightarrow [0,1],
\end{equation}

where $h$ denotes the hue value on the color wheel. These categories are not uniformly distributed along the axis, resulting in unequal perceptual coverage in the color space. To formally quantify this property, we introduce the \textit{Wideness} metric ($W_i$), which captures the extent of each hue category. Mathematically, it is defined as the measure of the $\alpha$-cut at level $\alpha=0.5$:

\begin{equation}
    W_i = \int_{0}^{360} \mathbb{I}\big(\mu_i(h) \ge 0.5\big) \, \mathrm{d}h
\end{equation}

where $\mathbb{I}(\cdot)$ is the indicator function. Geometrically, for convex functions such as the triangular and trapezoidal membership functions used in this model, $W_i$ represents the \textbf{length of the horizontal segment at the $y=0.5$ level}. The integral form is essential for maintaining consistency for the Red category, which is topologically split by the $0^\circ/360^\circ$ boundary. Table \ref{tab:hue_metrics_combined} shows the wideness values for each category from the COLIBRI model.

\begin{table*}[ht]
\centering
\caption{Hue Category Metric Analysis: Wideness and Boundary Widths (COLIBRI)}
\label{tab:hue_metrics_combined}
\begin{tabular}{lcccc}
\toprule
\textbf{Hue Category $H$} & \textbf{Wideness Range ($\alpha = 0.5$)} & \textbf{Wideness $W_i$ ($\alpha = 0.5$)} & \textbf{Left Boundary Width $\Omega_{i-1,i}$} & \textbf{Right  Boundary Width $\Omega_{i,i+1}$} \\ 
\midrule
Red       & $340.5^\circ - 12.5^\circ$  & 32.0 & 21.00 & 15.00 \\ 
Orange    & $12.5^\circ - 40.0^\circ$   & 27.5 & 15.00 & 12.00 \\ 
Yellow    & $40.0^\circ - 55.5^\circ$   & \textbf{15.5} & 12.00 & 19.00 \\ 
Green     & $55.5^\circ - 151.5^\circ$  & \textbf{96.0} & 19.00 & \textbf{47.00} \\ 
Cyan      & $151.5^\circ - 180.5^\circ$ & 29.0 & \textbf{47.00} & \textbf{11.00} \\ 
LightBlue & $180.5^\circ - 199.5^\circ$ & 19.0 & \textbf{11.00} & 27.00 \\ 
Blue      & $199.5^\circ - 255.0^\circ$ & 55.5 & 27.00 & 30.00 \\ 
Violet    & $255.0^\circ - 300.5^\circ$ & 45.5 & 30.00 & 45.00 \\ 
Magenta   & $300.5^\circ - 340.5^\circ$ & 40.0 & 45.00 & 21.00 \\ 
\bottomrule
\end{tabular}
\end{table*}

Adjacent categories share certain intervals, forming a shared Boundary Width $\Omega_{i,i+1}$, which represents the total extent of the transition zone between two linguistic terms. Formally, this width is defined as the measure of the intersection of the supports of two adjacent membership functions:\begin{equation}\Omega_{i,i+1} = \int_{0}^{360} \mathbb{I}\big(\mu_i(h) > 0 \wedge \mu_{i+1}(h) > 0\big) , \mathrm{d}h\end{equation}Geometrically, this operation corresponds to identifying the interval on the hue axis where both membership functions coexist with non-zero values. In the context of fuzzy sets, $\Omega_{i,i+1}$ represents the base of the intersection triangle (see Fig. \ref{fig:intersectionarea}), calculated as the distance between the point where $\mu_{i+1}$ begins to rise and the point where $\mu_i$ reaches zero.


This value quantitatively describes perceptual uncertainty: larger $\Omega$ values correspond to smooth, gradual transitions, while smaller values indicate sharp boundaries. We listed all boundary width values for each category in Table \ref{tab:hue_metrics_combined}.


The proposed analysis is fundamentally enabled by the COLIBRI models, which provide a perceptually grounded fuzzy representation of color categories. Unlike traditional color models such as RGB, HSV, or CIE Lab, which operate on fixed numerical values and lack explicit representation of category boundaries and overlap, COLIBRI models colors as distributions of membership across linguistic categories. This representation enables direct quantification of both category extent and boundary uncertainty using the proposed metrics. Such analysis is not feasible within conventional color models, as they do not capture the ambiguity and overlap inherent in human color perception.

\section{EXPERIMENT RESULTS AND ANALYSIS}

\begin{figure}[ht]
    \centering
    \includegraphics[width=\linewidth]{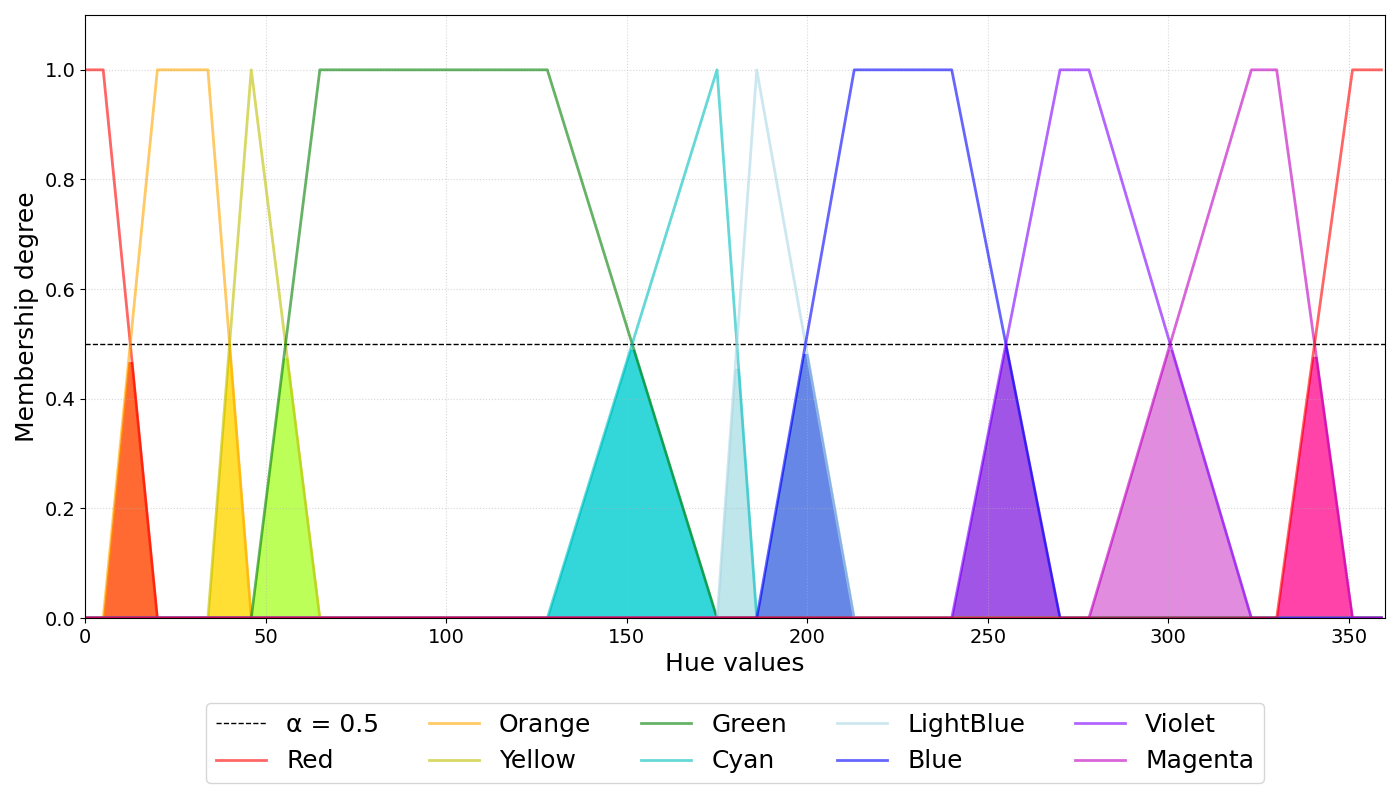}
    \caption{Fuzzy representation of hue categories with $\alpha$-cut ($\alpha = 0.5$). Differences in the extent and overlap of adjacent membership functions highlight perceptual asymmetries in color categorization.}
    \label{fig:intersectionarea}
\end{figure}

Figure~\ref{fig:intersectionarea} illustrates the fuzzy membership functions defined over the hue domain, along with the $\alpha$-cut at $\alpha = 0.5$. The intersections with this threshold define the boundaries between adjacent color categories.

Importantly, the category widths are highly non-uniform. The yellow category occupies a relatively narrow region of the hue space, whereas the green category spans a substantially wider interval. This imbalance is also reflected in the asymmetric overlaps between neighboring categories. Such differences indicate that color categories are not evenly distributed in perceptual space.

\begin{figure}[ht]
    \centering
    \includegraphics[width=\linewidth]{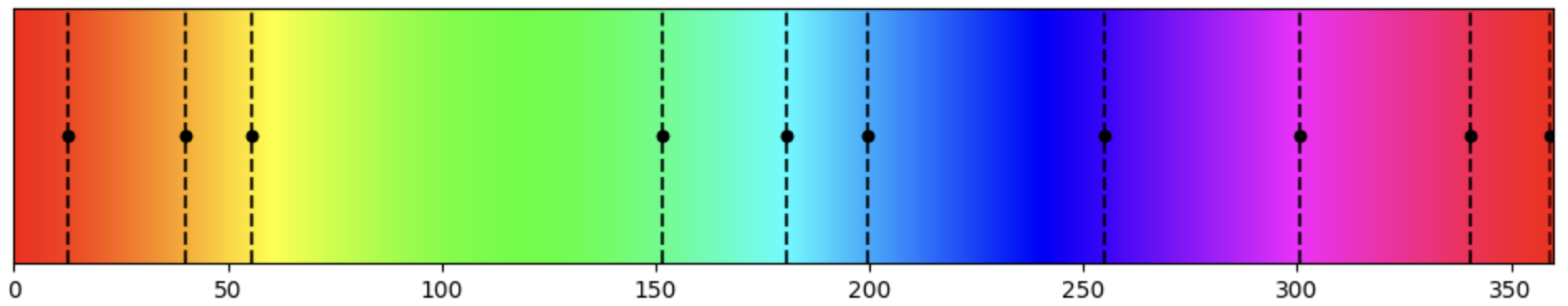}
     \caption{Hue spectrum with category boundaries defined by the $\alpha$-cut ($\alpha = 0.5$) of fuzzy color membership functions.}
    \label{fig:hue_boundaries}
\end{figure}

Figure~\ref{fig:hue_boundaries} shows the projection of the derived category boundaries onto the continuous hue spectrum. The boundary locations are obtained from the $\alpha$-cut ($\alpha = 0.5$) of adjacent fuzzy membership functions.

The resulting partition is clearly non-uniform. In particular, the interval corresponding to yellow is significantly narrower than that of green, which spans a much larger portion of the hue space. This asymmetry provides direct evidence that perceptual color categories differ in their extent, rather than forming evenly spaced segments.

Together, these results demonstrate that the asymmetry between yellow and green arises primarily from differences in category extent, with yellow forming a compact category and green covering a broader range of hues.

These findings further show that color categories differ not only in their location within the hue space but also in their structural properties. In particular, narrow categories such as yellow correspond to more precise, constrained perceptual labels, in which observers show greater agreement and sharper boundaries. In contrast, wider categories such as green cover a broader range of hues and exhibit more gradual transitions, indicating greater tolerance in human color naming.

\section{Discussion}

Across multiple lines of research, the yellow region on the color wheel is perceptually narrow, whereas green spans a wider swath of hue space. Behavioral and neural measures show that hue changes are faster and more sensitive near yellow than near green, meaning the yellow region covers a smaller angle or band of hues, while green spans a wider sector of the color circle \cite{Chiba2023A}, \cite{Macyczko2025Asymmetries}. 
Studies of categorical perception and discrimination similarly indicate that category boundaries around green and blue encompass larger ranges than around yellow \cite{EMERY201751, Raskin1983Perceptual}. Our findings corroborate theirs, although we used a different approach (fuzzy modeling). One explanation is that warm colors (red, yellow, orange) and cool colors (green, blue) are represented differently in the brain. Lightness affects warm colors more strongly, making their region behave differently from green/blue \cite{EMERY201751, Rosenthal2021Color}.

Another study reports that hue changes are perceived more rapidly, and categories are narrower, near yellow (and to some extent blue) than near green or magenta, which occupy broader ranges of color space
\cite{Macyczko2025Asymmetries}. In contrast, the magenta region of COLIBRI is modeled with a comparatively narrower support, indicating stronger perceptual differentiation than previously suggested.

Unlike prior perceptual studies, our model explicitly encodes hue asymmetry through parameterized fuzzy membership functions, allowing quantitative analysis of category width and transitions.


\section{Conclusion}


This paper presented a focused analysis of perceptual asymmetry in human color categorization using previously collected data from the COLIBRI experiment. By applying the proposed measures of category extent (Wideness) and boundary uncertainty (Boundary Width), we showed that hue categories are not only fuzzy, but also structurally non-uniform in their geometric organization.

The analysis revealed a clear asymmetry between yellow and green: the yellow category spans only 15.5$^\circ$ of hue space, whereas green covers 96.0$^\circ$, making green approximately 6.2 times wider. This asymmetry is also reflected in broader transition regions associated with green, indicating that perceptual color categories differ not only in their boundaries, but also in their extent. Together, these results show that uniform hue partitioning overlooks important perceptual structure and that category extent should be considered explicitly in perceptually grounded color modeling. The proposed Wideness and Boundary Width measures provide a quantitative basis for analyzing such asymmetries in future studies.

A limitation of this study is that the fuzzy partitions were derived from stimuli in which intensity and saturation were held constant, while only hue varied. As a result, the estimated category boundaries and category extents reflect perceptual organization under these controlled conditions and may shift under different intensity or saturation levels. Future work should examine the stability of the proposed metrics across broader variations in stimulus appearance.

\section*{Acknowledgement}
This research has been funded by the Science Committee of the Ministry of Science and Higher Education of the Republic of Kazakhstan (Grant No. AP22786412)


\bibliography{access}

\end{document}